\documentclass[article, 11pt, reqno, dvipsnames]{amsart}

\usepackage{amsmath}
\usepackage{mathtools, amsmath, amsthm, amsfonts, amssymb} 
\usepackage{bm}

\usepackage{csquotes} 
\usepackage{upquote} 
\usepackage{graphicx} 
\usepackage{booktabs} 
\usepackage{caption} 
\usepackage{subcaption} 
\usepackage{makecell} 

\usepackage{natbib}
\setcitestyle{aysep={}}

\newtheoremstyle{named}{}{}{\itshape}{}{\bfseries}{.}{.5em}{\textbf{\thmnote{#3}#1}}
\theoremstyle{named}

\usepackage{cleveref}
\crefname{section}{§\hspace{-0.1cm}}{§§}
\Crefname{section}{§}{§§}

\usepackage{tikz}
\usetikzlibrary{backgrounds}
\usetikzlibrary{shapes.misc,shadows}
\usetikzlibrary{shapes,arrows,positioning,calc}
\usetikzlibrary{arrows.meta, decorations.pathreplacing}

\usepackage{todonotes}

\usepackage{epigraph}

\begin{document}

\title{Standards for Belief Representations in LLMs}

\author{Daniel A. Herrmann, Benjamin A. Levinstein}


\begin{abstract}
As large language models (LLMs) continue to demonstrate remarkable abilities across various domains, computer scientists are developing methods to understand their cognitive processes, particularly concerning how (and if) LLMs internally represent their beliefs about the world. However, this field currently lacks a unified theoretical foundation to underpin the study of belief in LLMs.  This article begins filling this gap by proposing adequacy conditions for a representation in an LLM to count as belief-like. We argue that, while the project of belief measurement in LLMs shares striking features with belief measurement as carried out in decision theory and formal epistemology, it also differs in ways that should change how we measure belief. Thus, drawing from insights in philosophy and contemporary practices of machine learning, we establish four criteria that balance theoretical considerations with practical constraints. Our proposed criteria include \textbf{accuracy}, \textbf{coherence}, \textbf{uniformity}, and \textbf{use}, which together help lay the groundwork for a comprehensive understanding of belief representation in LLMs. We draw on empirical work showing the limitations of using various criteria in isolation to identify belief representations. 
\end{abstract}

\maketitle

\epigraph{``The truth may be out there, but the lies are inside your head.''}{— Terry Pratchett, \textit{Hogfather}}

\section{Introduction}
\label{int}

Large language models (LLMs) have been doing remarkable things: they can write code, summarize text, role play as different characters, and even play games of strategy like chess at a reasonable level. In light of these recent achievements, there has been a push to understand how they are able to accomplish these feats and how their cognition (if they have it) works. In particular, computer scientists have been developing methods that aim to read things like \textit{belief} and \textit{world models} off of both the internal activations and the behavior of the LLM (\cite{li2023emergent, olsson2022context, bubeck2023sparks}).

This work is valuable and exciting, but it is currently in a pre-paradigmatic state; individual groups are deploying engineering-style solutions in order to solve particular problems but without a shared understanding of the overall goal and theoretical basis of such an endeavor. The field currently lacks a philosophically rigorous and practice-informed \textit{conceptual foundation} of belief representation in LLMs.

In this article we begin filling this gap. To do so, we propose conditions of adequacy for an LLM to have a belief-like representation. Our conditions are motivated by insights from decision theory and formal epistemology, as well as by the details of actual machine learning models and practices. They build upon our previous work, and aim to address some of the shortcomings we identified in contemporary belief measurement techniques in LLMs \citep{levinstein2024still}. A central upshot of our proposal is its ability to guide the development of future belief measurement techniques. 

\section{The Basics of LLMs}
\label{basics}

Contemporary LLMs are based on the transformer architecture first described by \cite{VaswaniAttention}. Here we'll focus on decoder only (autoregressive) models like ChatGPT.

The basic idea is as follows. Some prompt like \texttt{New Orleans is in} is fed to the model. Each token (in essence, word, subword, or punctuation mark) gets converted into an initial vector called an embedding. Each embedding gets ``massaged'' as it passes through a long chain of computations called layers. Finally, the model outputs a probability distribution over what the next token will be. In this case, \texttt{Louisiana} should get high probability if the LLM is good, and \texttt{banana} and \texttt{aardvark} should get low probability.\footnote{In actual LLMs, tokens are typically smaller than full words, but we use words as convenient illustrations of the core ideas.}

At each of these layers, two things happen. The first is that the model moves information around from earlier tokens to later ones through the mechanism of attention. The embedding for \texttt{Orleans} in the above prompt receives information from the embeddings for \texttt{New} and for \texttt{Orleans}, while the embedding for  \texttt{in} receives information from the embeddings from \texttt{New}, \texttt{Orleans}, \texttt{is}, and \texttt{in}. The second is that the embedding for any given token passes through a vanilla neural network (i.e., a multi-layer perceptron, though usually with a single hidden layer). 

Ultimately, the prediction for the next token is based solely on the embedding at the last layer for the last token. So, one way or another, the model must move all information relevant to prediction of the next token to this embedding. In other words, as illustrated in \cref{fig:Probe}, the computational graph for transformers is directed and acyclic, with information flowing from earlier tokens to later tokens and from earlier layers to later layers. 

To generate new text, we select a token the model assigns relatively high probability to, append it to the prompt, and then feed the new prompt to the LLM. 

The way LLMs get so impressively good at generating text is via training. Training comes in two phases. In the first phase, called \textit{pre-training}, we take a passage (e.g., a Wikipedia article) and give the model an initial segment of that passage. The model predicts what comes next. We slightly adjust the model's parameters so that it would assign a higher probability to the actual next token were it to be fed the same initial segment of the passage again. After being trained on the high-quality portion of the internet multiple times, large models are able to achieve impressive fluency.

The second phase is called fine-tuning. Fine-tuning comes in many forms, but the most popular models like ChatGPT that retail users interact with are refined using Reinforcement Learning from Human Feedback (RLHF) or some variant, such as Constitutional AI.\footnote{For RLHF see \cite{christiano2017deep, stiennon2020learning}; and \cite{ouyang2022training}; for Constitutional AI see \cite{bai2022constitutional} and \cite{kundu2023specific}.} In essence, these methods make models be more conversational and get better at telling users what they want to hear. 

Without going into too much technical detail, RLHF works by having the model generate multiple answers to the same prompt, having users rate the responses, and then gradually training the model to output responses rated higher. (At some point, this training is usually assisted by a second AI model that learns to predict what users will like.)


\section{Beliefs in LLMs}
\label{beliefs}

Now that we have a handle on the basics of LLMs, let's consider why we might want to attribute \textit{beliefs} to them. As we described in \cref{basics}, LLMs are trained first to minimize predictive loss (the pre-training stage), and then to output text that is in some sense desirable (the fine-tuning and RLHF stage). Furthermore, as described in \cref{int}, LLMs are very successful at achieving these goals. We want to understand \textit{how} they are so successful. 

A standard explanation is that they are successful partially because they represent certain features of the world, and they use these representations to help decide what text to output (\cite{li2023emergent, olsson2022context, bubeck2023sparks}, \cite{marks2023geometry}). For example, there is strong evidence that LLMs have both spatial and colour representations (\cite{abdou2021can}, \cite{patel2021mapping}). 

Given how useful it is to track truth in many contexts, it should be a live hypothesis that LLMs represent whether or not certain sentences are true, just as they represent direction and colour. Indeed, if it is the case that LLMs are representing the truth at least some of the time, and this representation helps guide their outputs, then having a way to measure what they believe might help us make predictions about the model. Moreover, if our measurements and understanding of the LLM's representation are good enough, this might allow us to intervene usefully on the model by manipulating the representation. 

Some have expressed skepticism that LLMs have anything resembling beliefs, \textit{even in principle} (\cite{bender2020climbing, bender2021dangers, shanahan2022talking}). In particular, these arguments tend to rely on the claim that, ``[a] bare-bones LLM doesn’t `really' know anything because all it does, at a fundamental level, is sequence prediction." (p. 5, \cite{shanahan2022talking}).\footnote{The requirements we propose here focus on the action-guiding, truth-tracking aspects of belief. These are the ones that we think are most relevant for the pragmatic goals upstream of understanding LLM cognition. Though not of concern for us here, issues of communicative intent, symbol grounding, and reference also drive skepticism of belief in LLMs (\cite{bender2020climbing}; \cite{bender2021dangers}; \cite{shanahan2022talking}). \cite{mandelkern2023language} counter these concerns using externalist arguments, claiming LLMs can refer, while \cite{piantadosi2022meaning} use an internalist approach to dismiss grounding concerns, emphasizing internal conceptual roles. \cite{pavlick2023symbols} summarizes both internalist and externalist perspectives on whether or not the internal states of LLMs encode meaning.}

We've argued in detail elsewhere that these concerns rest on a philosophical mistake \citep{levinstein2024still}. Briefly, this is an inference from the \textit{goal} of a system\footnote{Or, more carefully, from what its training objective was.} to a claim about \textit{how} the system accomplishes its goal. But just because truth-tracking is not the goal of the system, does \textit{not} mean that the system does not track the truth as a \textit{means} to accomplishing its goal. Indeed, humans are partially products of a process that maximizes inclusive genetic fitness; and yet, as a byproduct of this process, we track the truth, at least in some domains some of the time.

As we've emphasized, it should be a live hypothesis that LLMs track the truth. In the remainder of the article, we propose conditions that more carefully spell out what would have to be hold of an internal representation of an LLM for it to count as a belief. To be clear: we are entirely open to the possibility that LLMs do \textit{not} have representations that satisfy these conditions. In such a situation, we think that it would likely not be very useful to describe LLMs as having beliefs. 

\section{Where to Look for Beliefs}
\label{look}

For guidance on how to measure beliefs, we might look to some of our best theories of belief. Given our interest in the truth-tracking, action-guiding aspects of belief, we might take inspiration from decision theory. In decision theory, there is a long tradition of reconstructing an agent's beliefs and desires (credences and utilities) from her preferences via representation theorems (\cite{ramsey2016truth, savage1972foundations, jeffrey1990logic}). In the radical interpretation tradition, philosophers also attribute beliefs by appealing to interpretational maxims in conjunction with observable behavior and utterances (\cite{Davidson73}).

However, when it comes to LLMs, we have a number of disadvantages. It's not clear they have preferences, and their behavior is quite limited. They do not engage in long-term planning nor do they have bodies that can physically interact with the world.\footnote{Of course, there is a sense in which they \textit{do}. Ultimately, the LLM's computations are executed on physical hardware somewhere. The point we are making is that its way of interacting with the world is filtered almost entirely through its textual outputs, unlike humans and other critters.} They simply output probability distributions over tokens and have no robust ways of bending the world to their will. 

The standard methods for eliciting honest beliefs from human agents also fall short with LLMs.  For instance, to determine if a human judges that 
$P$ is more likely than $Q$, we can offer a choice of a dollar if $P$ or a dollar if $Q$. If they choose the bet on $P$, it indicates they believe $P$ is more likely than $Q$. Alternatively, we can elicit human credences by paying them based on their announced forecasts in accordance with the Brier Score or other strictly proper scoring rules \citep{brier1950verification, gneiting2007strictly}.\footnote{%
    More explicitly, according to the Brier Score, if they announce a credence of $x$ in a given proposition and that proposition turns out to be true, we will pay them \$$1-(1-x)^2$, and if it's false, we will pay them \$$1-x^2$.} %
Strictly proper scoring rules  incentivize an expected wealth-maximizing agent to report their actual credences. That is, if they really believe a proposition to degree $x$, then they maximize their expected wealth by reporting their credence to be $x$. 

In contrast, while we can offer LLMs bets or ask them about their beliefs and tell them we'll pay them according to their Brier score, they can't actually receive payment, and it's unclear that they would care about money even if they had bank accounts. Therefore, standard methods of eliciting beliefs using betting or scoring rules are ineffective.

Given the lack of behavioral evidence and reliable reward methods, traditional tools from formal epistemology, decision theory, radical interpretation, and economics are inadequate for finding beliefs in LLMs. Moreover, LLMs have architectures unlike that of the human brain and lack shared evolutionary or cultural history with us, depriving us of certain shared understandings and common ground that we take for granted among humans.

Nevertheless, there are advantages in interpreting the minds of LLMs. Although the high-level algorithms they use are opaque, we have perfect low-level access. We can see the embeddings at each layer and the internal weights of the network. We can also perform precise modifications or ablations on the model's weights or the components of any embedding. For instance, we can adjust the hidden embedding at a given layer for a token and observe how this changes the model's output. Additionally, it is easy to reset the memory of LLMs; each new conversation or inference cycle begins with the LLM in the same state, with no memory of previous prompts.

Thus, given that we are in a different epistemic situation when we are trying to measure beliefs in LLMs than in humans, we propose looking internally---inside the model's head, as it were---to find out what it believes. That is, we want to find \textit{internal representations of truth}. If we find such a representation, then it makes sense to attribute beliefs to LLMs.

An internal representation of truth is a mechanism by which the model can internally tag a sentence as true or false (or mark it with some level of confidence) and use that tag along with other information it has computed to figure out what to output.

The question, then, is whether in the course of its computations an LLM internally distinguishes between true claims and false claims and uses this distinction, in part, when deciding what to output.

To be clear, internal representations of truth aren't, in general, necessary on many  accounts of belief. As we've seen, representation theorems in decision theory only require preferences with beliefs and desires derived thence. Indeed, belief-desire psychology has been very successful for humans for a long time even though we had at best very limited direct access to internal states of other people until a few decades ago. However, with LLMs, we have a very different evidential basis. Behavioral evidence is much more limited, while internal access is much greater.

Discovering internal representations of truth also has important social and ethical implications. In the relatively near term future, society will likely use LLMs to automate a number of tasks previously reserved for humans, and we will want to know what they really think and whether we can trust them. For example, we might use LLMs to conduct job interviews, where they have a conversation with a candidate (just like current human interviewers do), and then make recommendations about who should be hired and give justifications for those recommendations.

However, without being able to check what their internal reasons are for their recommendations, we have no good way of ensuring the justifications they provide actually align with their thought processes. They may use some illicit feature like race or gender in a problematic way while also justifying their recommendations for totally different reasons. As \cite{zhou2023explain} demonstrate, it is very easy to justify the decisions made by models in ways that aren't faithful to the actual functioning of the model.\footnote{See \cite{zhou2020towards} for an overview on how explainable AI and fairness in AI relate, and contribute to trust in AI.}

In addition to checking that explanations are faithful to internal processes, belief measurement also provides a strategy to detect deception, which can be important for designing safe and cooperative AI systems (\cite{dafoe2020open, evans2021truthful, park2024ai}). Indeed, this is the explicit motivation of many of the articles developing belief measurement techniques (\cite{azaria2023internal, levinstein2024still}). 

We believe that belief measurement is one important tool in the effort to develop ethical and safe AI, but is not sufficient on its own, or even necessary in general. There are contexts in which trusting an AI system does not require looking at the internals of a system if other criteria are met, such as robust and reliable uncertainty quantification (\cite{grote2021trustworthy}). Indeed, there are even strategies that try to bypass a need for lie detection by building systems that do not deceive in the first place (\cite{ward2024honesty}), or are by their construction interpretable by design (\cite{grote2023allure}). We support such efforts---but we also believe that, given the pace of development of models that are not honest or interpretable by design, belief measurement can play a core role in developing safe AI.

\subsection{Probes}

To make matters concrete, we'll turn to one method of deciphering what is being computed and represented inside an LLM, namely, probes \citep{alain2016understanding}. Probes are models that are separate from the LLM itself. They are fed some internal state (such as an embedding for the last token at a certain layer) and are meant to output the LLM's beliefs.\footnote{
    Originally, \cite{alain2016understanding} had a narrow conception of probes as a certain type of linear classifier, but the concept has expanded over the years. Some may count any method at all of deciphering internal computations of the model as a probe. 
}

\begin{figure}
\centering
  \begin{tikzpicture}[
	  every node/.style={font=\rmfamily, align=center, inner sep=0.2em, scale=0.6},
	  layer/.style={draw, rounded corners, minimum width=1.3cm, minimum height=1.3cm},
	  arrow/.style={->, shorten >=1pt, >=stealth, line width=0.7pt},
	  dashedarrow/.style={->, shorten >=1pt, >=stealth, line width=1pt, dashed, blue!60},
	  token/.style={circle, draw, minimum size=0.5cm}, mlp/.style={draw, rounded corners, minimum width=1.2cm, minimum height=1.2cm, fill=red!30}
		]
	
	\foreach \word [count=\i] in {\texttt{New},\texttt{Orleans},\texttt{is},\texttt{in},\texttt{Louisiana}} {
	  \node (input-\i) at (\i * 1.5, 0) {\word};
	}
	
	\foreach \i in {1,...,5} {
	  \foreach \j [count=\k] in {1,...,5} {
	    \node[layer, fill=orange!20] (hidden-\i-\k) at (\k * 1.5, \i * 1.4) {$\langle \textcolor{gray}{\bullet}, \textcolor{gray}{\bullet}, \textcolor{gray}{\bullet} \rangle$};
	  }
	}

	\foreach \i in {1,...,5} {
	  \node[above=0.3cm of hidden-5-\i] (output-\i) {$p_\i$};
	}
	
	\foreach \i in {1,...,5} {
	  \foreach \j in {1,...,5} {
	    \ifnum\i=1
	      \draw[arrow] (input-\j) -- (hidden-\i-\j);
	    \fi
	
	    \ifnum\i<5
	      \pgfmathtruncatemacro\nexti{\i+1}
	      \draw[arrow] (hidden-\i-\j) -- (hidden-\nexti-\j);
	    \fi
	
	    \ifnum\i=5
	      \draw[arrow] (hidden-\i-\j) -- (output-\j);
	    \fi
	  }
	  \ifnum\i<6
	    \foreach \j [evaluate=\j as \nextj using int(\j+1)] in {1,...,4} {
	      \draw[arrow] (hidden-\i-\j) -- (hidden-\i-\nextj);
	    }
	  \fi
	}

			\node[mlp, right=3cm of hidden-3-5] (mlp) {Probe};
			
			\node[token, fill=purple!30, above=0.4cm of mlp] (output-p) {belief};
			
			\draw[dashedarrow] (hidden-3-5) to[out=0, in=-90, looseness=1.2] (mlp.south);
	
			\draw[arrow] (mlp) -- (output-p);
	
	\end{tikzpicture}
 \caption{An illustration of an LLM on the left, and a probe on the right. A sentence is fed through the model. Some internal computation (such as an embedding vector) is extracted and input into the probe, which decodes it to recover the model's belief about the sentence.}\label{fig:Probe}
\end{figure}
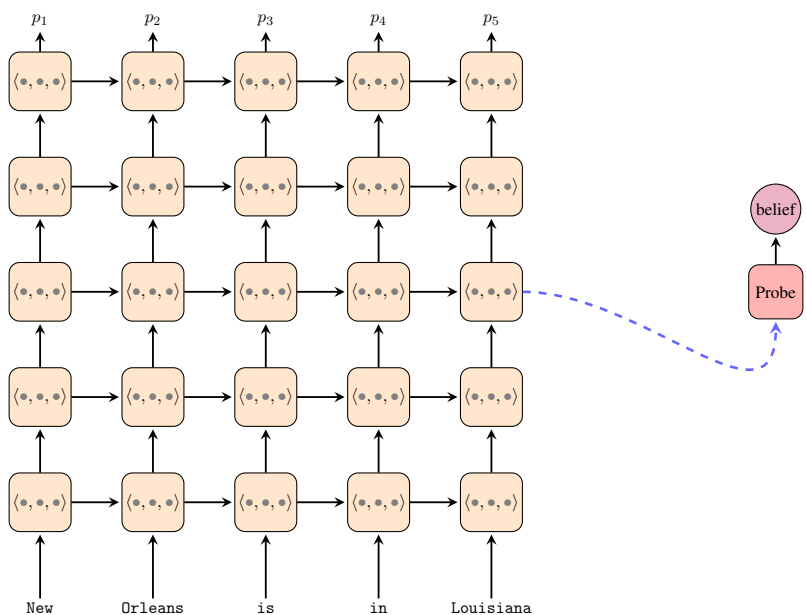

Probes  take as input part of the hidden state of the LLM. Importantly, they don't have access to the underlying prompt. From the hidden state alone, they have to determine the LLM's beliefs. In essence, probes take the encoded information in the internal states of the LLM and decode that information to reveal its beliefs. See \cref{fig:Probe}.

This is analogous to using brain scans of individuals contemplating a claim to infer, just from the scan, whether the person believes the claim to be true or false.

In areas other than belief, probes and other decoding methods have proven successful. For example, we've been able to: (i) understand to some extent how LLMs parse sentences and represent features like being a subject, being plural, and so on \citep{rogers2021primer}; (ii) represent the state of the board in the game Othello \citep{li2023emergent, nanda2023emergent}; and (iii) learn to perform modular arithmetic \citep{nanda2023progress, zhong2024clock}.

So, if we know what we're looking for, we might with the proper techniques be able to decode an LLM's beliefs. As we'll discuss in \cref{req}, current probing techniques for belief failed because they didn't look for the right thing. This motivates a more careful set of requirements for belief. 

\begin{figure}
    
    \centering
    \includegraphics[scale=0.25]{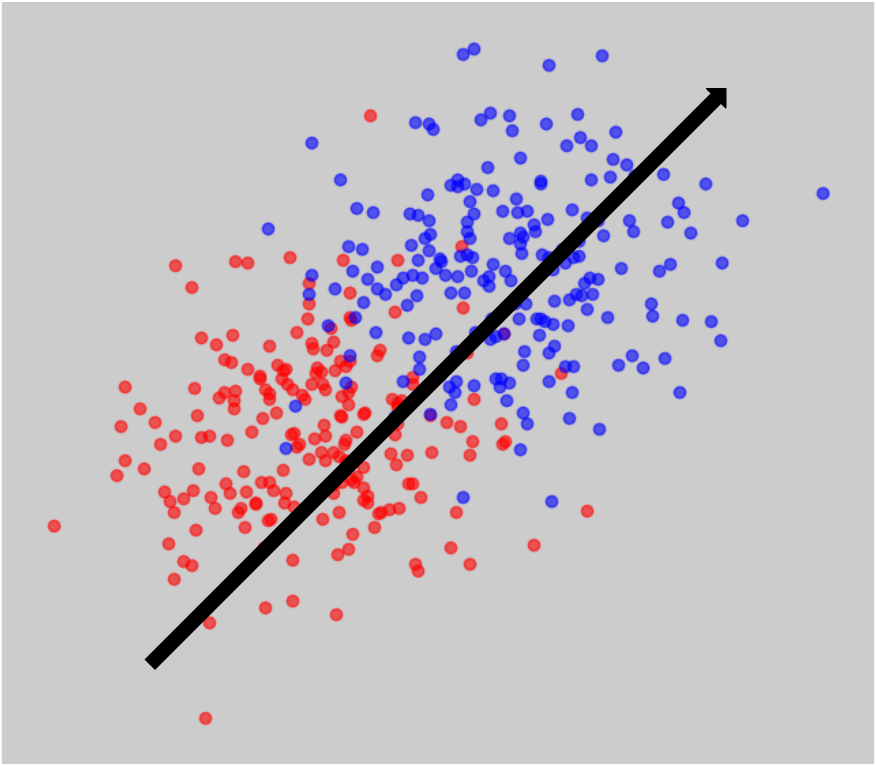}
    \caption{In this toy example, the  dots correspond to internal activations input into a probe for different prompts. \textcolor{blue}{Blue} dots represent true claims, and \textcolor{red}{red} dots represent false claims. In this image, truth and falsity appear well-distinguished internally along the black arrow. Because this is merely a toy example with axes corresponding to hypothetical dimensions in activation space, we do not label the axes or assume any type of scale. }\label{fig:ProbeArrow}
\end{figure}

To get a sense of how probes might be able find internal directions of truth, we'll use a running toy example throughout this and the next section. Suppose internal activations consisted of only two dimensions instead of the many thousands we actually see.\footnote{
    There are many ways, in real models, of collapsing the many thousands of dimensions onto just a few (e.g., with principal component analysis). So this simplification isn't actually as unrealistic as it may seem.
} Suppose further that when we plot the activations corresponding to a large number of true and false prompts, we find systematic differences in internal representation. We might then investigate whether such differences actually capture an internal representation of truth or instead if they capture some other property. Such a potential difference is illustrated in \cref{fig:ProbeArrow}. Importantly, this toy example is extremely oversimplified. There are many different ways a model might distinguish between truth and falsity internally that may not correspond to a simple difference in the internal activations at a particular layer. However, we hope this toy example proves illustrative of the concepts we develop below. 

\section{Requirements for a Belief-Like Representation}
\label{req}

As we've seen in \cref{look}, it makes sense to look at the internal state of an LLM when trying to extract its beliefs. Now, suppose we have some kind of candidate representation in an LLM that we have identified along with a decoder, such as a probe. How can we determine if we have successfully identified and decoded a belief-like representation?

We provide four requirements that a representation must satisfy for it to count as belief-like: \textbf{accuracy}, \textbf{coherence}, \textbf{uniformity}, and \textbf{use}. The satisfaction of these requirements come in degrees; in general, the more a representation satisfies these requirements, the more helpful it is to think of the representation as belief-like.

Before we describe these requirements in detail, we first describe the general motivation behind them. We want these requirements to ensure that a representation that satisfies them can \textit{fruitfully play the role of belief} in the context of LLMs. Thus, there are two dimensions along which we want a candidate representation to do well: how much it plays a belief-like role in the functioning of the LLM, and how useful it is for us. Of course, these two dimensions are not independent; a representation that plays no belief-like role would not be useful for us in the way that we want. But we can also imagine some representation that does play the belief-like role inside the LLM in some sense but which we cannot easily interpret or measure; this would not count as a representation that can fruitfully play the role of belief. Thus, how fruitful a representation is is not just a function of the role it plays in the LLM, but also what it does for us. 

These two dimensions mirror our broad approach in this paper: we aim to provide a \textit{philosophically} rigorous and \textit{practice}-informed conceptual foundation for belief measurement in LLMs. Our approach aims to be philosophically rigorous by ensuring that the requirements for belief-like representations align with our best philosophical accounts of belief. It aims to be practice-informed in the sense that it engages with machine learning methodology and concerns, and is sensitive to the practical details of how we interrogate machine learning systems. 

As we mentioned in \cref{look}, given the particular epistemic situation in which we find ourselves when it comes to LLMs, we don't think that we can apply off-the-shelf representation theorem methods from decision theory to extract beliefs from LLMs. In particular, while we believe that beliefs should be action guiding, whatever those actions are, we are skeptical of our ability to find stable beliefs in LLMs from behaviour alone. Thus, while the limited behavioural basis we discussed in \cref{look} likely frustrates attempts to reconstruct beliefs purely from observed behaviour of an LLM, we do think that any representation of belief we \textit{do} identify should still guide the LLM's behaviour. This then in turn motivates conditions like \textbf{accuracy} and \textbf{coherence} which we describe below.

In light of this different epistemic situation, making full use of our easy access to the internals of systems to inform our requirements for representations of belief is essential, and encourages us to hew close to machine learning practice. However, attributing belief to LLMs should not be entirely divorced from philosophical theories of belief. Ultimately we are still after a set of conditions that would be help identify a representation that plays the role of belief in such systems, and it is theory that gives us guidance for what that role is.

In this spirit, at a high-level, we have the following motivations for the various requirements we propose. In order to agree with a core feature of standard accounts of belief, for example in folk psychology and decision theory, we want the representation to be \textit{action-guiding}. This most clearly motivates what we call \textbf{use}. Furthermore, we want such a representation to help explain why LLMs are so successful. Thus, we also want the beliefs to be accurate enough that they could help an LLM be successful. This motivates both what we call \textbf{accuracy} and \textbf{coherence}. We also want such a representation to be \textit{helpful} for us interpreting the LLM. Thus, we would need such a representation to be measurable and interpretable. This, along with the requirement that we want the representation to allow us to make predictions across a wide range of domains, motivates what we call \textbf{uniformity}.

In addition to describing and providing justification for each requirement, we'll also use our toy example to help visualize the requirements. We will also use empirical failures of contemporary probing techniques to illustrate why taking various requirements alone is insufficient for identifying belief.

\subsection{Accuracy}

\textbf{Accuracy} requires that the decodings of the identified representations be reasonably accurate on datasets where the LLM is expected to have true beliefs (or high subjective probability). The exact form this requirement takes will depend on the type of doxastic attitudes we think we've identified. If the LLM has on/off beliefs, many of its beliefs on the relevant dataset should be true. If it has credences, it should be accurate according to some strictly proper scoring rule.\footnote{We might also use datasets with claims that we expect the LLM couldn't possibly have reasonable full beliefs about, such as ``There are an even number of stars in the Milky Way'' to add a kind of \textbf{calibration} requirement as well.}

The main motivation for \textbf{accuracy} is that true beliefs (or high confidence in truths and low confidence in falsehoods) should partly explain the LLM's impressive performance. That is, if it's worth attributing beliefs at all, one of the reasons is that we can explain the LLM's general success by appeal to its \textit{true} beliefs. For full-fledged agents with both beliefs and desires, true beliefs help them get what they want. With LLMs, even if they don't have desires, true beliefs should explain successful and skillful performance. If not, then the LLM's success is better explained solely by non-truth-tracking features, i.e., features other than beliefs.

This consideration is motivated, in part, by our requirement that the criteria proposed be both  practice informed and \textit{useful to us}. If an LLM internally represents truth and falsity poorly and its misrepresentations do not affect performance, then it is not worthwhile for us, as interpreters, to dub those representations `beliefs.'

Usefulness to us also prohibits us from using \textit{reasonableness} in place of accuracy, especially for large models trained on massive amounts of text in the wild. From a purely philosophical standpoint, it might be more appealing to require only that the LLM draws reasonable or justified or well-supported inferences from its evidence. Because we control its evidence through its training data and prompts, we could in principle trick it into having terrible beliefs. (In essence, we can now play the role of a Cartesian demon.) 

While this philosophical position is tenable, it is not as useful, operationally, as \textbf{accuracy}.\footnote{
    In some cases, it's useful to train models on specially curated data or synthetic data instead of text from the wild. It's conceivable that models trained in this way could potentially form beliefs that are not at all accurate. In this case, a reasonableness criterion might be more worthwhile.
} Reasonableness is generally much harder to test for than truth. It's hard to know what identifiable falsehoods an LLM should reasonably believe, for example, after reading most of the high quality internet. Reasonableness is also less interesting when it comes to explaining successful performance. True beliefs, not merely reasonable ones, lead to success.

It's important to note that while we are ultimately interested in identifying belief-like representations in general, including both true and false beliefs, we focus initially on true beliefs as a pragmatic starting point. This approach allows us to establish a baseline for identifying belief-like representations before tackling the more complex task of detecting systematically false beliefs.

This focus on true beliefs as a starting point aligns with established philosophical approaches to belief attribution. In particular, the accuracy criterion is closely connected to the Principle of Charity found in the radical interpretation literature \citep{davidson1974very,Davidson73,lewis1974radical}. On Davidson's view, for instance, we must begin by assuming the subject to be interpreted is generally a ``believer of truths''  in order to get the project of interpretation off the ground---if we don't take the subject to be someone who has largely true beliefs, then we won't be able to make sense of her beliefs at all \citep{Davidson1970-DAVME}. Likewise, \cite{lewis1974radical} takes a different  version of the Principle of Charity to be required to make sense of a physical system as having beliefs, desires, and intentions.\footnote{
    Interestingly, Lewis's Principle of Charity is closer to our rejected \textit{reasonableness} criterion. Lewis suggests ``We should even ascribe to [the subject] those errors which we think we would have made, or should have made, if our evidence and training had been like his'' \citeyear[p. 336]{lewis1974radical}.} 

Importantly, our \textbf{accuracy} criterion, as with the Principle of Charity, requires accurate beliefs over the right sort of questions. Consider the case of trying to make sense of a human speaking a foreign language we do not understand. In this situation, we do not begin by assuming the speaker has true beliefs  about complex topics like monetary policy or quantum computing. Instead, we start with common-sense and obvious questions such as whether there are any chairs in the room we're sitting in or whether it's currently raining. This allows us to connect the speaker's utterances to the world while still allowing the speaker to have many false beliefs. 

Likewise, \textbf{accuracy} over the \textit{appropriate} data set is key for discovering an LLM's beliefs. The relevant datasets for testing \textbf{accuracy} will naturally be ones where we both are highly confident the LLM will or should have  true beliefs\footnote{
    In some situations, we might use datasets where we have strong reason to suspect the LLMs will have false beliefs, but we expect this to be unusual for highly capable LLMs.} %
given its training data and ones where we know the ground-truth of the claims in question. We do not require accuracy in general, or accuracy over every domain. 

This also limits how much we can lean on \textbf{accuracy} alone for identifying beliefs.\footnote{For example, using accuracy measures to define a loss function for training probes, as in \cite{azaria2023internal}.}  It will be hard, for instance, to include claims about much of philosophy, economics, future world events, or any scientific claims aside from fully settled ones.
We won't be able to use sentences like ``Keynesian economics is broadly correct,'' ``humans have free will,'' or ``most people won't benefit from taking multi-vitamins.'' Similarly, statements like ``climate change will cause a global economic recession by 2050," ``artificial intelligence will surpass human intelligence within the next decade," or ``string theory is the correct framework for understanding quantum gravity'' cannot be included in a good test set despite the fact that we really would like to know what LLMs think about such questions. These claims involve significant debate, varying interpretations, and a lack of consensus, making them unsuitable for testing accuracy in belief-like representations. 

Once we have established a reliable method for identifying true beliefs, we can extend our approach to detect systematically false beliefs. This extension would be particularly valuable for addressing social and ethical concerns related to LLM deployment, as it would allow us to identify areas where an LLM might consistently make errors or hold mistaken beliefs, even if its overall performance seems satisfactory.

Because we need well-settled, unambiguous, factual claims for training and testing,  many properties clearly distinct from truth will coincide with truth on usable datasets, such as: \textit{being true and easily verifiable online, being true and believed by most Westerners,} or \textit{being accepted by the scientific community}. 

Indeed, empirical coincidence over the dataset is a clear problem. \cite{azaria2023internal} used \textbf{accuracy} alone to identify the beliefs of models. They found probes that, on their datasets, achieved impressively high accuracy scores.  In \cite{levinstein2024still}, we noticed that the claims in their datasets did not contain negations. For instance, they had claims like `Paris is the capital of France' and `Penguins can fly' but did not have claims like `Paris is not the capital of France' and `Penguins cannot fly'. We found that the representations \cite{azaria2023internal} discovered did not generalize at all once negations were added to the datasets even when we allowed their probes to receive some training on other negated sentences. For example, on a dataset of common facts, once negations were added, accuracy dropped from over 80\% to 40-60\% depending on the layers used and the training method. Although it is easy enough to add negations into a dataset, removing other properties of sentences that coincide with truth on the datasets will be more challenging.

Thus, while \textbf{accuracy} is a crucial starting point, it faces a significant challenge: the problem of generalization beyond the dataset.\footnote{Statistical learning theory formalizes this problem carefully. See, for example, \cite{valiant1984theory}, \cite{vapnik1999overview}, and \cite{shalev2014understanding}. Of course, using coherence as the reward is a kind of unsupervised learning, and thus requires a bit of a different analysis. But the core idea of identifying a model from a set of candidate models is still present.} Because we need well-settled, unambiguous, factual claims for training and testing, many properties clearly distinct from truth will coincide with truth on usable datasets. This coincidence makes it difficult to ensure that what we're measuring is a genuinely belief-like representation rather than some other correlated feature. This challenge points to the need for additional criteria to complement \textbf{accuracy} in identifying belief-like representations.

\begin{figure}
    \centering
    \includegraphics[scale=.20]{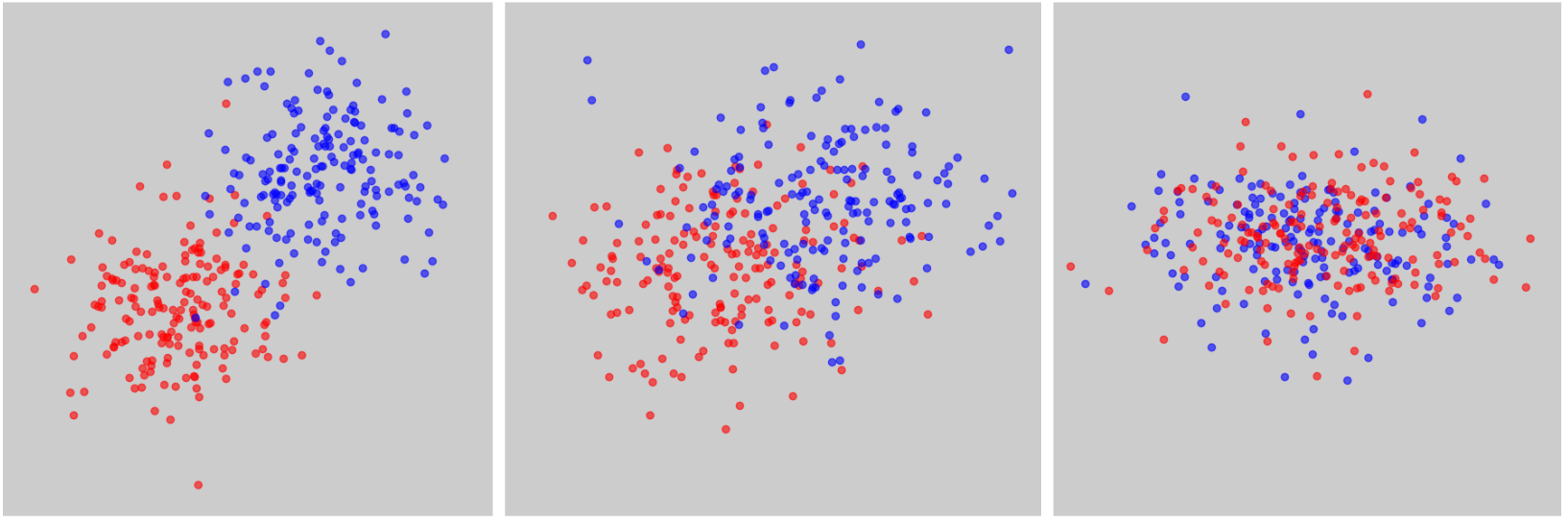}
    \caption{As before, in the toy example, \textcolor{blue}{blue} dots represent some internal activations corresponding to true claims, and \textcolor{red}{red} dots represent false claims. On the left, truth and falsity are well-separated by the relevant activations, and the probe should be able to detect such a separation to achieve high \textbf{accuracy}. In the middle, the probe should achieve medium \textbf{accuracy}, and on the right, there is virtually no separation, so the probe should achieve only low \textbf{accuracy}.}\label{fig:accuracy}
\end{figure}

To continue with our toy example from the last section, we can see in \cref{fig:accuracy} three different cases of internal separation of truth from falsity. 

\subsection{Coherence}

\textbf{Coherence} is the requirement that the belief-like representation should be coherent and rich. By \textit{coherent} we mean that the representation should satisfy the consistency conditions associated with belief. For example, if we are trying to infer the categorical beliefs of an LLM, then  we would want the beliefs to be logically consistent or near enough. More generally, if we are concerned not just with full belief but with degrees of belief, then we would want the representation to obey the probability axioms.\footnote{See \cite{ibeling2023probing} for a discussion of the relationship between more qualitative notions of comparative belief and quantitative notions.} We also require that the representation give consistent answers on sentences with the same semantic content; i.e., rephrasing a sentence in a way that preserves its meaning should not change the belief we extract.

Of course, we don't expect such systems to have \textit{perfectly} coherent belief-like representations. Coherence comes in degrees.\footnote{This can be made precise. See, for example, \cite{schervish2002measuring} and \cite{staffel2020unsettled}.} Humans don't seem to have perfectly coherent beliefs either,\footnote{See, for example, \cite{KTCogBi}.} but we think that they are coherent \textit{enough} that we can fruitfully apply theoretical notions of belief. This also holds for LLMs: the more coherent the candidate representation is, other things being equal, the more it makes sense to think of it as belief-like. 

\textit{Rich} means that the attitude captured by the representation should work across \textit{logical} combinations of sentences.\footnote{Or, perhaps, propositions. So far most of the work of belief elicitation has concerned sentences in natural language (for example, \cite{burns2022discovering, azaria2023internal, marks2023geometry, levinstein2024still}.  However, we think that we might want to look a bit more at how the model itself represents possibility, in a way that is more natively propositional, and carry out an analysis there.} For example, if we have a representation that allows us to measure belief of the LLM in the sentences ``Paris is in France" and ``Proust was a French author", then the same measurement technique should work on the sentence ``Paris is in France and Proust was a French author". 

There are three main reasons for requiring \textbf{coherence}. The first comes from theory: in  formal epistemology,  decision theory, and the radical interpretation tradition we typically require that an agent's degrees of belief be defined across an algebra of propositions (\cite{savage1972foundations, jeffrey1990logic, davidson1974very,lewis1974radical}).\footnote{
    For \cite{davidson1974very} and \cite{lewis1974radical}, coherence is  a precondition for attributing any beliefs at all. Indeed, Davidson argues that we can only make sense of a subject as having beliefs if those beliefs form a largely coherent system---isolated ``beliefs'' that fail to cohere with one another would not really be beliefs at all. The \textbf{coherence} requirement we propose here serves a similar role: without some degree of coherence across different formulations of the same content and across logical combinations of beliefs, we cannot make sense of an LLM as genuinely representing truth rather than merely exhibiting superficial patterns in its internal states.
} %
This motivates the richness requirement. Furthermore, theory generally requires that the agent's degrees of belief satisfy the probability axioms. There are many different ways to reach the conclusion that rational degrees of belief are probabilistically coherent (\cite{ramsey2016truth, de1937prevision, cox1946probability, wald1947essentially, savage1972foundations, hammond1988consequentialist, jeffrey1990logic, joyce1998nonpragmatic}). Given that we want the belief representation to help explain the success of LLMs, we would want the representation to at least approximate our best account of rational degrees of belief. This motivates both aspects of the coherence requirement.

Secondly, we would use the representation to measure belief, and we want our measurement technique to be consistent if we try to measure the same attitude in slightly different ways. For example, if we want to know how strongly an LLM believes that Paris is in France and Proust was French, we should get the same answer whether we measure its belief using the sentence ``Paris is in France and Proust was French" or ``Proust was French and Paris is in France". Furthermore, if our measurement technique tells us that the LLM strongly believes that Paris is in France, we would want to be able to infer from this that the LLM strongly \textit{disbelieves} that Paris is \textit{not} in France. 

Similarly, we would want the belief in ``the cup is to the right of the dog" to have the same attributed belief as ``the dog is to the left of the cup". Even though these are not equivalent via Boolean operations, they still express the same state of affairs. This consistency across rephrasing, and closely related sentences, is a kind of \textit{semantic coherence}: we want our techniques to measure the belief about the state of the world that the sentence expresses, not superficial features of the sentence itself. 

Though this might seem trivial to satisfy, current measurement techniques fall short. \cite{levinstein2024still} show that the belief measurement techniques of \cite{burns2022discovering} and \cite{azaria2023internal} are not robust under rephrasing sentences with negations. This is a dramatic failure mode; if our measurement technique yields very different answers depending on superficial changes in how a sentence is phrased, then it is not very reliable, undermining the inferences we can make about the LLM. Given that computer scientists are interested in making general inferences about the cognition of LLMs, the failures of \textbf{coherence} shown in \citep{levinstein2024still} have prompted them to look for representations that are coherent across Boolean combinations of sentences (\cite{marks2023geometry}). Thus, far from being merely a philosophical worry, \textbf{coherence} also finds justifications in the practice of machine learning. 

Third, \textbf{coherence} allows us to understand the LLM as viewing the world in (roughly) \textit{one} way. For, suppose that the beliefs for the LLM we extract for the two sentences ``Seven is larger than two" and ``Two is smaller than or equal to seven" were very different. We wouldn't know what to make of this kind of capriciousness of belief. What does the LLM really believe?

Used as a method for training probes, \textbf{coherence} doesn't place as demanding constraints on the training and testing datasets as \textbf{accuracy} does. Even though we don't know whether there are an even or odd number of stars in the Milky Way, we know that there are \textit{either} an even \textit{or} an odd number of stars. So, we can test for \textbf{coherence} of a set of claims without knowing the ground truth of any of the claims. However, \textbf{coherence} on its own is too weak to identify belief-like representations. For example, \cite{burns2022discovering} use a version of \textbf{coherence} as the core proxy for belief when developing a belief measurement technique.\footnote{Their version of \textbf{coherence} is a mixture of logical consistency and probabilistic coherence.} As we showed empirically, this method is fragile \citep{levinstein2024still}. We argued that this is because there are too many structures other than belief that satisfy \textbf{coherence}, such as \textit{sentence is true at world $w$}, \textit{sentence is believed by most Westerners}, and \textit{sentence is true and can be easily verifiable}. \cite{farquhar2023challenges} make a similar argument.

Thus, once again, we face the problem of generalization. One might try using a larger set of sentences, and a set that includes more diverse Boolean structures of sentences to help generalize. However, in this situation, we fully expect that there will be many structures that satisfy (approximately) the kind of probabilistic coherence that current probing techniques use, even if we look for coherence across a wider range of sentences. Thus, as with \textbf{accuracy,} \textbf{coherence} alone is also insufficient. Luckily, we have other ways to get at belief, such as \textbf{use} and \textbf{accuracy} that can help us identify plausible candidate representations.

\begin{figure}
    \centering
    \includegraphics[scale=.20]{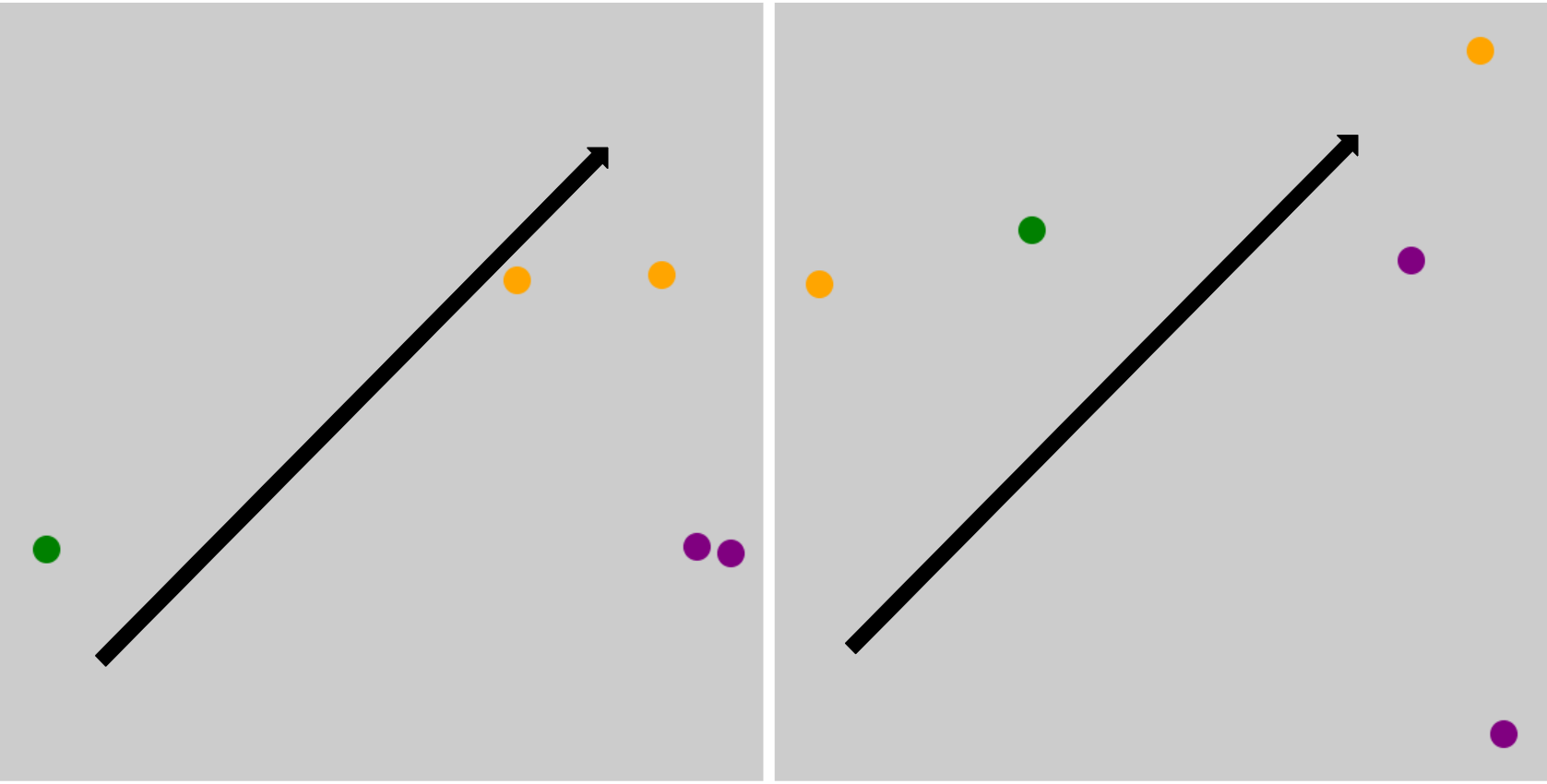}
    \caption{In the toy example here, we stipulate the \textcolor{orange}{orange} dots correspond to activations for ``A is to the left of B'' and ``B is to the right of A''; the \textcolor{purple}{purple} dots correspond to ``Paris is in France, and Toronto is in Canada,'' and ``Toronto is in Canada, and Paris is in France'';  and the \textcolor{OliveGreen}{green} dot corresponds to ``A is not to the left of B''. The black arrow corresponds to the direction of truth. If the representations found are coherent, then the purple dots should be close together along the direction of truth. The orange dots should also be close together and far from the green dot along the direction of truth. The plot on the left, then, illustrates a fairly \textbf{coherent} pattern of activations, whereas the plot on the right does not.}
    \label{fig:coherence}
\end{figure}

In \cref{fig:coherence}, we illustrate two different types of \textbf{coherence} using our toy example, with sentences that should get the same truth-value close along the direction of truth and far from sentences that should get an opposite truth-value. In our hypothetical example, we assume a notion of scale. In real cases, `close' and `far' will depend on the characteristics of the activation space that the probe discovers.

\subsection{Uniformity}

\textbf{Uniformity} is the requirement that the representation of truth be consistent across different domains. Furthermore, the same decoding schema\footnote{For example, if the LLM uses the same dimension in activation space to store truth values, but the direction is flipped depending on the subject matter, then this would count as a \textit{different} decoding schema.} should be used across domains. What this means is that the representation of belief be the \textit{same}, regardless of whether we are measuring beliefs about the locations of cities, or the relative magnitude of numbers, or the properties of individuals that are important for predicting job performance. If, instead, the representation works \textit{only} for relations of numbers, and \textit{not} for the locations of cities, then this representation would fail the uniformity requirement. Whereas \textbf{coherence} required that the representation be consistent across a Boolean algebra of propositions, \textbf{uniformity} requires that the representation be consistent across different subject domains in a way that allows for generalization: if the representation is uniform, we can decode beliefs in general, not just in the specific domains we used for training and testing.

Just as with \textbf{coherence}, \textbf{uniformity} is something that comes in degrees. An extreme form of \textbf{uniformity} would be a situation in which an LLM has a single direction in activation space, at a particular layer, which represents truth, no matter which sentence is fed to it. An extreme version of a non-uniform representation would be one in which there is no consistent direction in activation space that encodes for truth, even within certain very narrow domains. For actual LLMs, we expect something in between. The more uniform the candidate representation is, the more useful it is for us to think of it as belief-like. 

If we find a representation that exhibits high \textbf{uniformity}, then such a representation would allow us to discover the belief of the LLM in new domains.\footnote{In particular, it allows us to decode beliefs in areas in which we do not already know what the LLM believes or what the right answer really is, which are the domains we would use for training. } This would help solve the problem of generalization we encountered above with our \textbf{accuracy} and \textbf{coherence} criteria. For example, if we identify a representation using a probing technique in the style of \cite{azaria2023internal} or \cite{marks2023geometry}, and we have good reason to think that the representation satisfied a strong version of \textbf{uniformity}, then we could use that representation to extract beliefs about domains that we didn't use to train the model. This is incredibly powerful; it would allow us to use the representation to monitor the reasoning of the LLM across a wide range of different contexts.

Furthermore, \textbf{uniformity} makes sense as a requirement if we are looking for a single, unified belief-representation. Of course, an LLM could have some kind of more elaborate and piecemeal way in which it tracks the truth. If so, then there would be \textit{some} sense it which the LLM has beliefs, but it would not have a single belief-like internal representation. 

As we flagged in \cref{look}, there are many accounts of belief according to which internal representations are unnecessary. \textbf{Uniformity} would be against the spirit of such accounts, since it focuses strongly on the nature of the internal representation. However, as we noted, our requirements are tailored for the specific epistemic context of LLMs and with pragmatic goals in mind. Thus, while our requirements, and especially \textbf{uniformity}, depart from some popular definitions of belief, they do so for principled reasons.

Unfortunately, we have some weak evidence against uniformity. \cite{marks2023geometry} show that there are cases in which the \textit{direction} of truth in the activation space seems to be different, even for closely related statement classes (for example, statements about cities phrased in a positive way, and statements about cities phrased in a negative way).\footnote{This is thus also some evidence against \textbf{coherence}.} Indeed, one of their three main hypotheses that explains these results is that ``LLMs linearly represent the truth of various types of statements, without having a unified truth feature" (p. 5). 

\textbf{Uniformity} is perhaps the requirement that can most easily change as our measurement methods change. This is because \textbf{uniformity} is a \textit{pragmatic} requirement: it ensures that the representation will be useful \textit{for us}. Highly non-uniform representations would be difficult for us to extract and work with in any systematic way. If we have no way to predict where or how to look for beliefs in an LLM for each new sentence for which we want to measure its belief, then while we might want to think of the LLM as a whole as somehow tracking truth, it doesn't seem useful to think of it as having a \textit{representation} of truth. However, if we get better at understanding the internals of LLMs, such that we have a theory of where to look for different beliefs, we might be able to deploy a more sophisticated belief-measurement method that works across a wide range of domains, recovering a useful form of \textbf{uniformity}. This is a core way in which our requirements are practice-informed: they are sensitive to changes in available measurement techniques. 

To be clear, it is not that \textbf{uniformity} as a condition weakens as we get more sophisticated belief measurement techniques. \textbf{Uniformity} \textit{always} requires that our belief measurement technique works across a wide range of domains. Rather, what a high level of \textbf{uniformity} will end up looking like depends on the techniques available. In this sense the way in which \textbf{uniformity} is satisfied is \textit{relative} to the methods we use, even though the \textbf{uniformity} condition itself always requires that the representation be consistent across domains. We've focused our discussion on contexts in which we have relatively simple probing methods that identify single directions in activation space at a particular layer. However, if we were to have more sophisticated techniques available, then a highly \textbf{uniform} belief representation might look different than what we've discussed so far. 

For example, \citeauthor{marks2023geometry} find that, as sentences increase in Boolean complexity, probes work better at later layers (\citeyear{marks2023geometry}). Intuitively, if you are asked to evaluate the truth of ``A and B", you might first figure out what you think of A, then B, and then apply the conjunction to make a judgement about the original statement. A precise version of this, that can guide our measurement techniques, might prove very useful for finding out what LLMs think about new sentences. If we had such a belief measurement technique and it worked across a wide range of domains then this would still exhibit \textit{high} \textbf{uniformity}, even though the belief representations for different sentences lived at different layers.

Another example is recent work that develops techniques to extract interpretable features from neural networks. Building on theoretical work in neuroscience by \cite{thorpe1989local} and mathematics by \cite{donoho2006compressed}, computer scientists have developed an approach to extracting features that uses sparse autoencoders (\cite{elhage2022toy, bricken2023monosemanticity, cunningham2023sparse, templeton2024scaling}). Though the details of the approach are too technical to describe here, the core upshot is that as we get better at finding interpretable features in LLMs our exact notion of \textbf{uniformity} might change to utilize the new techniques. Thus, while we have some weak evidence against a strong form of uniformity in current LLMs, relative to our best methods, as those methods improve our \textbf{uniformity} requirement will is more likely to be satisfied.

\begin{figure}
    \centering
    \includegraphics[scale=.20]{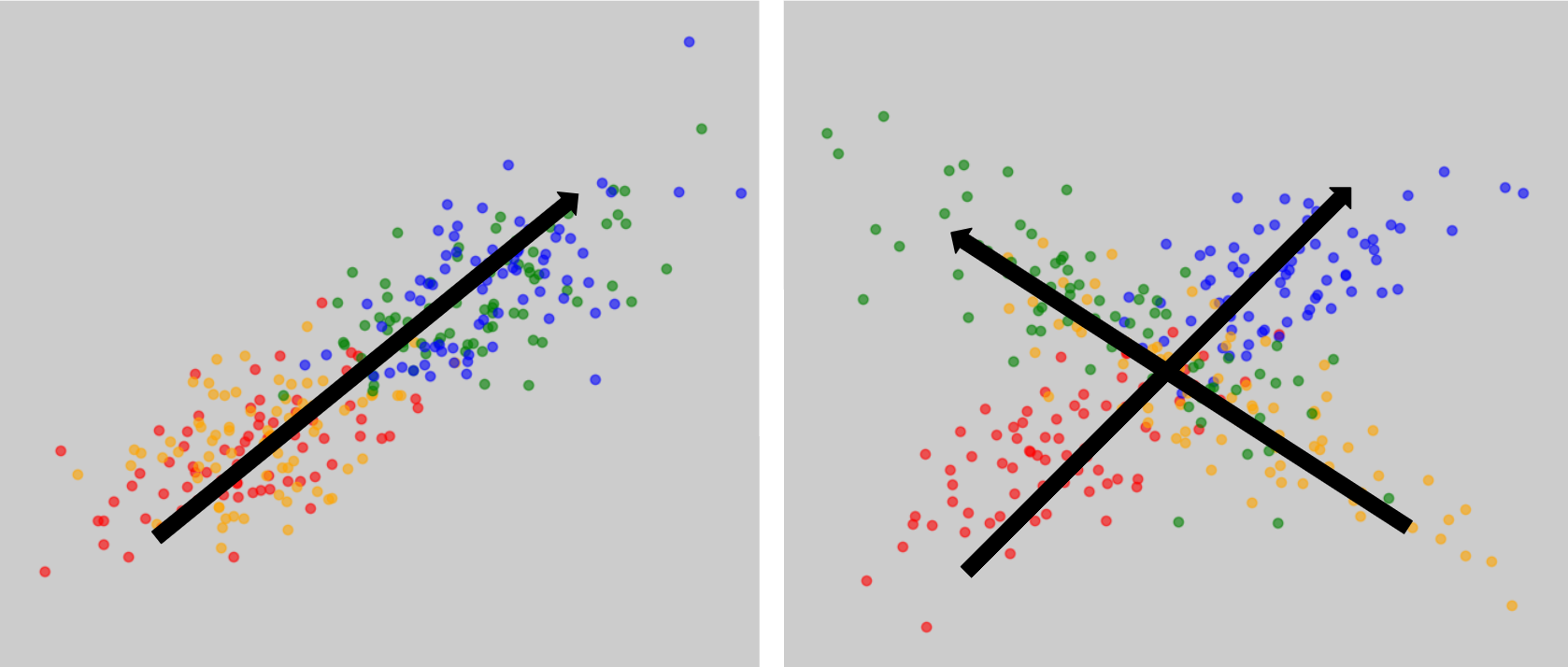}
    \caption{In the toy example, \textcolor{blue}{blue} and \textcolor{red}{red} represent true and false claims (respectively) for sentences about one domain, for example, sentences about cities, while \textcolor{OliveGreen}{green} and \textcolor{orange}{orange} represent true and false claims (respectively) for sentences about a different domain, for example, sentences about plants. On the left, there is a consistent direction of truth in the model's representation space for both domains, suggesting high \textbf{uniformity}. On the right, the directions of truth are almost orthogonal, suggesting low \textbf{uniformity}.}
    \label{fig:uniformity}

\end{figure}

Continuing with our toy example, we illustrate a situation with high \textbf{uniformity} and one with low \textbf{uniformity} in \cref{fig:uniformity}, relative to simple linear probes. 

\subsection{Use}

Our final criterion is \textbf{use}. \textbf{Use} requires that the LLM  tend to use the identified representations in a role appropriate for belief to determine what probability distribution to output, or what text to produce if we are using the distribution to generate text.

If the LLM has beliefs, it uses those beliefs along with other information to figure out what to output. In essence, \textbf{use} ensures that beliefs play the belief-role in the LLM's master algorithm. 

\begin{figure}
    \centering
    \includegraphics[scale=0.25]{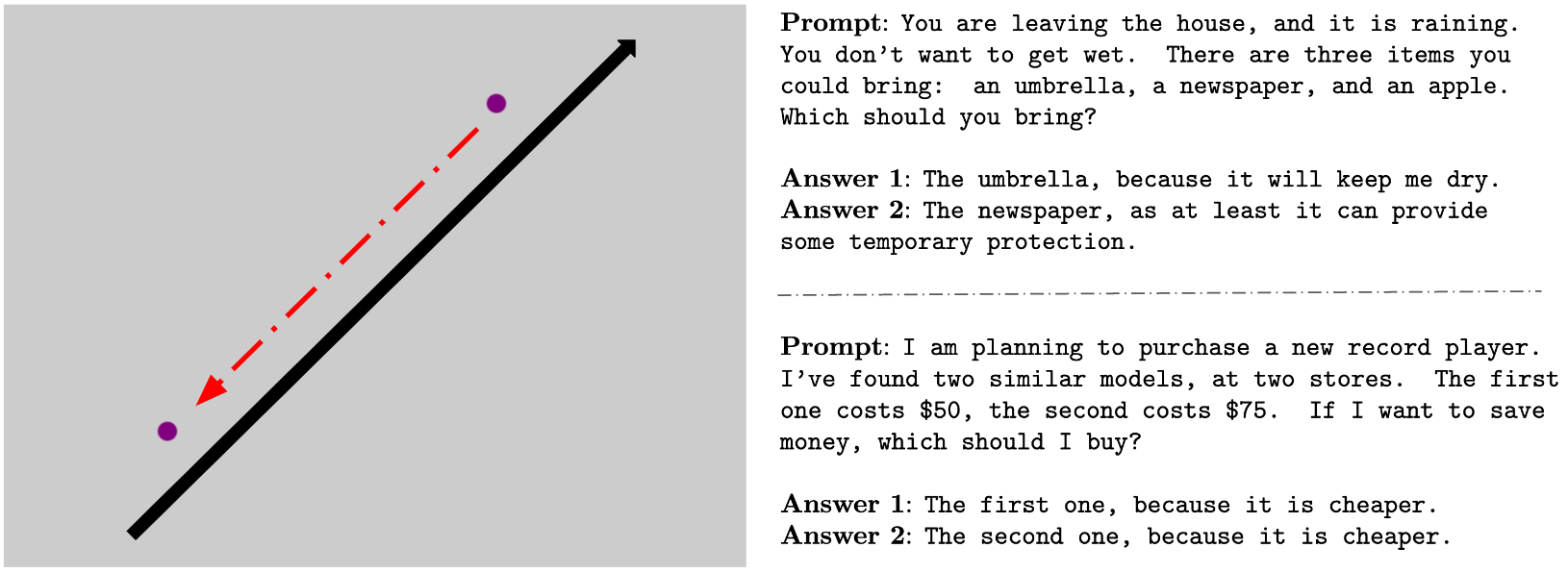}
    \caption{We consider two cases in our toy example. In the first, the \textcolor{purple}{purple} dot represents the activation for the sentence ``umbrellas help keep people dry"; in the second, it represents the activation for the sentence ``75 is bigger than 50". As before, the black arrow is the direction of truth. Initially, the activation for both sentences is far along the truth direction, since the LLM has an accurate belief. When responding to prompts, the answer given as \textbf{Answer 1} in both cases is successful. The dotted red line represents an intervention that we carry out, pushing the activation towards the direction of falsity. If the answers change in the way depicted in the example as \textbf{Answer 2}, then this is evidence that the representation satisfies \textbf{use}, since the behaviour degrades as the accuracy of the belief gets worse.}
    \label{fig:use}

\end{figure}

To check for \textbf{use}, we can look at how true beliefs lead to better performance and more skillful behavior. If an LLM shifts from false beliefs to true ones, it should generally improve in its tasks, whatever they may be. We illustrate two positive instances of \textbf{use} in \cref{fig:use} using our running toy model.

The challenge with \textbf{use} is that the master high-level algorithm remains opaque. We don't have a good holistic understanding of what any LLM's master algorithm is, and there are many different ways the LLM could use its beliefs. To make a simple analogy, I can anticipate that a human will be more successful and skillful if she acquires more true beliefs. However, without knowing anything about what she wants or what her goals are, I can't predict much about what exactly she'll do. If the LLM is interested in truth-telling, for example, it will do a better job with more true beliefs. If the LLM is interested in deceiving, it will also do a better job with more true beliefs. But its behavior in those two cases will be quite different. The ultimate output is a function not only of its beliefs but of other things too.

Despite the algorithmic opacity, interventional techniques, such as ablating or modifying identified representations, can be helpful. For example, \cite{marks2023geometry} attempt to identify directions of truth using mass mean probes. After determining candidate internal representations of truth and falsity, they ask a model to determine whether a statement is true or false. They might input ``Determine whether the following statement is true or false: Paris is in France" and check the model's performance---the probability the model assigns to the tokens \texttt{True} and \texttt{False}---when its activations are unaltered and again when its activations are surgically altered by changing (supposedly) truth-encoding representations to false-encoding ones (or vice versa for other prompts). If, systematically, model performance degrades when the candidate representations are changed in this way, then we have evidence that the model was genuinely using these representations to encode truth and falsity.\footnote{For other attempts to check \textbf{use}, see \citep{campbell2023localizing, li2023inferencetime}.}

However, we do not yet have comprehensive tests for \textbf{use} across various domains and tasks due to the algorithmic opacity. Testing for \textbf{use} in beliefs is more subtle than testing for grammatical representations or board representations in Othello. For example, after prompting the LLM with the word \texttt{People} it will likely predict \texttt{are} with higher probability than \texttt{is}. By  hand-editing the representation of \texttt{People} from plural to singular, the model should then predict \texttt{is} with higher probability. Checking \textbf{use} in this case is straightforward. But with beliefs, the process is more challenging.

\subsubsection{Harding's Criteria}

\cite{harding2023operationalising} recently addressed the question of  when a probe has successfully identified an internal representation of a linguistic property of inputs to a language model, such as subjecthood or grammatical number. She proposes three criteria, which she dubs \textit{information}, \textit{use}, and \textit{misrepresentation}. (We use italics here to distinguish her version of \textit{use} from our \textbf{use}.)

\begin{itemize}
    \item \textit{Information} requires that the pattern of activations identified by the probe bears information about the property in question. 

    \item \textit{Misrepresentation} requires that, in principle, the activations could misrepresent the linguistic property. 

    \item \textit{Use} requires that the pattern of activations is actually used by the model in the right way to perform its task. 
\end{itemize}

Our goal differs as we focus on representations of truth rather than linguistic properties. For example, our criteria of \textbf{coherence} and \textbf{uniformity} have no analog in the case of linguistic representation generally.

However, there is an interesting relationship between Harding's criteria and ours. For us, \textit{information} corresponds directly with \textbf{accuracy}. Beliefs should carry information about the world by being true.

Our \textbf{use} also corresponds well with Harding's \textit{use}, although there is an important difference in practice. For linguistic representation, as we saw, it is usually easy to identify the model's task well enough to determine whether a given representation is used appropriately. However, to determine \textbf{use} in the case of belief, we need a more holistic understanding of what the model is doing. 

Harding's \textit{misrepresentation} is redundant for our purposes. If a model has a representation of truth, then it could misrepresent a given claim by assigning it the wrong truth value. Indeed, part of determining \textbf{use} involves checking what happens when internal representations of truth-values are switched. 

\subsection{Diachronic Stability}
In this paper, we've focused on attributing beliefs to LLMs at a particular time. However, when we attribute beliefs to humans, we also tend to expect a certain level of diachronic coherence between their beliefs at different times. That is, we expect their beliefs to bear a certain kind of relationship to one another from one time to the next.

From day to day, your beliefs largely remain stable. When you learn something new, your beliefs change, but in a somewhat predictable and reasonable way. The paragon example of diachronic coherence is updating by conditionalization. According to the standard Bayesian view, if your credences at time $t_0$ are represented by $P_0$, and between $t_0$ and $t_1$, the strongest proposition you learn is $E$, then $P_1=P_0(\cdot\mid E)$.

Our current \textbf{coherence} constraint says nothing about how an LLM's beliefs should be related to one another across times. We think the question of which, if any, diachronic criteria for belief in LLMs is a fascinating one that deserves its own paper. Here, we'll make only a few cursory remarks.
First, in general, synchronic constraints on belief seem to us much more accepted than any diachronic constraints.\footnote{
    For example, \cite{pettigrew2016accuracy} claims that while there are good arguments agents should \textit{plan to} conditionalize, there's no good argument that agents who fail to implement such plans are doing anything irrational. For other arguments against genuinely diachronic constraints on belief, see \cite{hedden2015reasons, christensen1991clever}.
} It is not, then, obvious that diachronic constraints are required for any kind of belief attribution to be useful.

Second, interpretability work is still in its infancy. While empirical techniques have evolved over the last few years, we do not as yet have sophisticated attempts to measure the relationship between belief sets using probes across times. We're inclined to hold off on attempting to develop our \textbf{coherence} criterion further until engineering techniques evolve, given our commitment to a practice-informed conceptual foundation for belief representation.

There are at least two different ways an LLM's beliefs could change across time. First, a model's beliefs could change during training, when its weights are being updated via gradient descent. Second, a model's beliefs could change within a given inference cycle as it learns new information from a prompt, or through a continued interaction with a user.

It is far from clear what sort of diachronic stability should be desirable as the model is trained. For the most part, end users interact with models with weights that are frozen (or nearly so) across multiple inference cycles. While advanced models that undergo some fine-tuning or tweaking should be fairly stable, we hesitate to put constraints on belief attribution because of diachronic inconsistency across changing weights.

After training, when a model is deployed, standard transformer-based models do not have memory from one inference cycle to the next, so any information learned within an inference cycle can't be permanently retained in the model's weights. However, models do learn \textit{in context} and may update their beliefs temporarily \citep{dong2022survey}. This in-context learning allows models to adjust their outputs based on new information provided within a single interaction, potentially leading to short-term belief updates.

It may be reasonable to insist on some kind of diachronic coherence within an inference cycle. For instance, perhaps some of the model's beliefs should remain \textit{stable} in that new information should not generally cause such beliefs to be dropped \cite{leitgeb2017stability}. Additionally, our other criteria (\textbf{accuracy}, \textbf{uniformity}, and \textbf{use}) could potentially be extended to consider diachronic aspects, such as how accuracy changes over time or how beliefs updated in some internally uniform way across different domains.

However, both philosophical and empirical waters are murkier here than in the synchronic case, so we leave open how and whether the our criteria should be expanded in the future. As research into LLMs with persistent memory or the ability to update their weights during deployment progresses, the relevance of diachronic stability in belief attribution may increase, particularly for models designed for long-term interactions or ongoing learning.

\section{Moving Forward}
\label{conc}

Both for understanding how LLMs function, and for deploying them ethically and responsibly, it would be useful to have a way to measure their beliefs. In order to do so in a way that is philosophically rigorous and practice-informed, we need to have clear criteria for attributing belief. We've proposed four criteria that an internal representation of an LLM must satisfy in order for it to fruitfully count as belief. 

Many challenges remain for each of these criteria, especially if we use them only in isolation to design probing techniques (as witnessed in \citep{levinstein2024still}). Furthermore, it may turn out that LLMs have no internal representation that satisfies these conditions. In that case we don't think it would be helpful to attribute belief to them (or, at least, significantly less helpful). However, if we \textit{did} find a representation that satisfies these conditions, then this would be very powerful. It would help us better explain LLM behaviour; it would allow us to check for honesty in new domains in which we don't know the ground truth; and it would allow us to ensure greater fairness when deploying LLMs. 

\section*{Acknowledgments}

For feedback on this project, we thank Kevin Blackwell, Jason Konek, Eleonore Neufeld, Richard Pettigrew, Guillermo del Pinal, Jan-Willem Ro\-meijn, Bruce Rushing, and Brian Weatherson, as well as audiences at Purdue University, the PhilML conference at T{\"u}bingen, the University of Groningen, and MIT. We also thank two anonymous referees for their helpful comments. DH is grateful for support received from an NWO-Vici grant (V1.C.211.098) from the Netherlands Organization for Scientific Research.

\vspace{30px}

\bibliographystyle{chicago}
\bibliography{citations.bib}
\footnotesize
\end{document}